\documentstyle[colap]{article}

\newcommand{\dothanks}{\thanks{\ This work was sponsored by Teilprojekt
    B4 ``From Constraints to Rules: Compilation of {\sc hpsg}'' of the
    Sonderforschungsbereich 340 of the Deutsche
    Forschungsgemeinschaft.  I would also like to thank Dale Gerdemann
    and Guido Minnen for helpful comments on the ideas presented here.
    All remaining errors are of course my own.}}

\author{John Griffith\dothanks\\
  Seminar f\"ur Sprachwissenschaft\\ Universit\"at
  T\"ubingen\\ Kl. Wilhelmstr. 113,\\ D-72074 T\"ubingen, Germany\\
    griffith@sfs.nphil.uni-tuebingen.de}

\title{\vspace{-0.5in}Modularizing Contexted Constraints}

\def\p{^\prime}
\def\pp{^{\prime\prime}}

\newcommand{\tuple}[1]{\mbox{$\langle{#1}\rangle$}}

\newcommand{\dep}[2]{\tuple{_{#1}\: #2}}

\newcommand{\ol}[1]{\overline{#1}}

\newcommand{\bp}[1]{(#1)}

\def\iff{i\!f\!\!f}

\newcommand{\Case}{case}

\newcommand{\conf}{con\!f}
\newcommand{\dnf}{dn\!f}
\newcommand{\DNF}{D\!N\!F}

\newcommand{\card}[1]{|#1|}

\newcommand{\qed}{\hfill \(\Box\)}

\newcommand{\avmsz}{\tiny}

\def\avmarraystretch{1.2}

\newcommand{\idx}[1]{\mbox{\fbox{\tiny #1}\hspace{0.1em}}}

\newcommand{\avmlist}[1]{\mbox{$\langle #1 \rangle$}}

\newcommand{\avmu}[1]{%
  \renewcommand{\arraystretch}{\avmarraystretch}%
  \avmsz%
  \mbox{%
    $\left[%
    \begin{tabular}{@{}l@{\hspace{0.3em}}l@{}}%
      #1
    \end{tabular}
  \right]$}}

\newsavebox{\groupbox}
\newsavebox{\avmbox}
\newsavebox{\AVMbox}

\newcommand{\avmd}[2]{%
  \renewcommand{\arraystretch}{\avmarraystretch}%
  \avmsz%
  \sbox{\groupbox}{\makebox[0em][l]{\raisebox{-1ex}{${_{#1}}$}}}%
  \sbox{\avmbox}{\begin{tabular}{@{}l@{}}#2\\[-3ex]\end{tabular}}%
  \sbox{\AVMbox}{%
    $\left\{%
    \begin{tabular}{@{}l@{\hspace{.3em}}l@{}}%
      & \usebox{\avmbox}\\
      \usebox{\groupbox}%
      \end{tabular}
    \right\}$}%
  \mbox{%
    \begin{tabular}{@{}l@{}}%
      \\[-0ex]
      \usebox{\AVMbox}\\
      \\[-0ex]
    \end{tabular}
    }}

\newtheorem{definition}{Definition}
\newtheorem{lemma}{Lemma}
\newtheorem{theorem}{Theorem}

\newenvironment{Def}[1]{
  \begin{definition}[#1]\ \\[+.5\baselineskip]
    }{
  \end{definition}}

\newenvironment{Lem}[1]{
  \begin{lemma}[#1]\ \\[+.5\baselineskip]
    }{
  \end{lemma}}

\newenvironment{Thm}[1]{
  \begin{theorem}[#1]\ \\[+.5\baselineskip]
    }{
  \end{theorem}}

\begin{document}
\maketitle


\begin{abstract}
  This paper describes a method for compiling a constraint-based
  grammar into a potentially more efficient form for processing.  This
  method takes dependent disjunctions within a constraint formula and
  factors them into non-interacting groups whenever possible by
  determining their independence.  When a group of dependent
  disjunctions is split into smaller groups, an exponential amount of
  redundant information is reduced.  At runtime, this means that an
  exponential amount of processing can be saved as well.  Since the
  performance of an algorithm for processing constraints with
  dependent disjunctions is highly determined by its input, the
  transformation presented in this paper should prove beneficial for
  all such algorithms.
\end{abstract}

\bibliographystyle{acl}

\section{Introduction}
\label{sec:intro}
\newcommand{\NP}{N\!P}

There are two facts that conspire to make the treatment of disjunction
an important consideration when building a natural language processing
(NLP) system.  The first fact is that natural languages are full of
ambiguities, and in a grammar many of these ambiguities are described
by disjunctions. The second fact is that the introduction of
disjunction into a grammar causes processing time to increase
exponentially in the number of disjuncts.  This means that a nearly
linear-time operation, such as unification of purely conjunctive
feature structures, becomes an exponential-time problem as soon as
disjunctions are included.\footnote{Assuming $P \neq \NP$.}
Since disjunction is unlikely to disappear from natural language grammars,
controlling its form can save exponential amounts of time.

This paper introduces an efficient normal form for processing
dependent disjunctive constraints and an operation for compilation
into this normal form.  This operation, {\em modularization}, can
reduce exponential amounts of redundant information in a grammar and
can consequently save corresponding amounts of processing time.  While
this operation is general enough to be applied to a wide variety of
constraint systems, it was originally designed to optimize processing
of dependent disjunctions in feature structure-based grammars.  In
particular, modular feature structures are more efficient for
unification than non-modular ones.  Since in many current NLP systems,
a significant amount of time is spent performing unification,
optimizing feature structures for unification should increase the
performance of these systems.

Many algorithms for efficient unification of feature structures with
dependent disjunctions have been proposed
\cite{MK89,ED90,Ger91,Str92,Gri96}.  However, all of these algorithms
suffer from a common problem: their performance is highly determined
by their inputs.  All of these algorithms will perform at their best
when their dependent disjunctions interact as little as possible, but
if all of the disjunctions interact, then these algorithms may perform
redundant computations.  The need for efficient inputs has been noted
in the literature\footnote{Cf. \cite{MK91} for instance.} but there
have been few attempts to automatically optimize grammars for
disjunctive unification algorithms.

The modularization algorithm presented in this paper takes existing
dependent disjunctions and splits them into independent groups by
determining which disjunctions really interact.  Independent groups of
disjunctions can be processed separately during unification rather
than having to try every combination of one group with every
combination of every other group.

This paper is organized as follows: Section~\ref{sec:depIntro} gives
an informal introduction to dependent disjunctions and shows how
redundant interactions between groups of disjunctions can be reduced.
Section~\ref{sec:cxcons} shows how normal disjunctions can be replaced
by contexted constraints.  Section~\ref{sec:deps} then shows how these
contexted constraints can encode dependent disjunctions.
Section~\ref{sec:modn} presents the modularization algorithm for
contexted constraints.  However, even though this algorithm is a
compile-time operation, it itself has exponential complexity, so
making it more efficient should also be a concern.  A theorem will
then be presented in section~\ref{sec:freeCombElim} that permits an
exponential part of the modularization algorithm to be replaced by
combinatorial analysis.

\section{Dependent disjunctions}
\label{sec:depIntro}

Dependent disjunctions are like normal disjunctions except that every
disjunction has a name, and the disjuncts of disjunctions with the
same name must be chosen in sync.  For example, \(\dep{d}{\phi,
  \phi\p, \phi\pp} \wedge \dep{d}{\psi, \psi\p, \psi\pp}\) is a
conjunction of two dependent disjunctions with the same name, \(d\).
What this means is that if the second disjunct in the first
disjunction, \(\phi\p\), is chosen, then the second disjunct of the
other disjunction, \(\psi\p\), must be chosen as well.  (Note that
what kind of constraints the \(\phi\)s and \(\psi\)s are is not
important here.)  The computational reason for using dependent
disjunctions over normal disjunctions is that dependent disjunctions
allow for more compact and efficient structures.  This is particularly
true when dependent disjunctions are embedded inside of feature
structures.  This is because in that case disjunctions can be kept
local in a directed graph structure thus saving redundant feature
paths.

We say that disjunctions with the same name are in the same {\em
  group}.  One distinguishing feature of a group of disjunctions is
that each disjunction must have the same number of disjuncts.  This is
essentially where redundant interactions originate.  For instance, in
\(\dep{d}{\phi, \phi, \phi\p, \phi\p} \wedge \dep{d}{\psi, \psi\p,
  \psi, \psi\p}\) each disjunction has four disjuncts, but really only
two values.  But more importantly, no matter what value of the first
disjunction is chosen (\(\phi\) or \(\phi\p\)) the same values are
possible for the second (\(\psi\) or \(\psi\p\)).  In other words,
these disjunctions are actually independent from one another, and can
be put into different groups: \(\dep{d\p}{\phi, \phi\p} \wedge
\dep{d\pp}{\psi, \psi\p}\). This is the process of modularization
which will be formalized in section~\ref{sec:modn}.

One might be tempted to think that modularization is unnecessary since
grammar writers are unlikely to write dependent disjunctions which
contain independent parts.  However, grammar writers may not be the
only source of dependent disjunctions.  Many grammar processing
systems use high-level descriptions which are then transformed into
more explicit lower-level grammars.  This transformation process may
very well introduce large numbers of dependent disjunctions with
exactly this property.

One example of where this can happen is in the compilation of lexical
rules in \cite{Meurers:Minnen:95b}.  In this paper, Meurers and Minnen
describe a compiler which translates a set of {\sc HPSG} lexical rules
and their interaction into definite relations used to constrain
lexical entries.  In \cite{Meurers:Minnen:96}, they show how an
off-line compilation technique called constraint propagation can be
used to improve the definite clause encoding produced by their
compiler to allow for more efficient processing.  The use of dependent
disjunctions provides an attractive alternative to the constraint
propagation approach by specifying all the information associated with
a lexical entry directly as a single dependent feature structure
rather than hidden in a set of definite clauses.\footnote{In the case
  of infinite lexica, definite clauses are still necessary to encode
  recursive information.} Consider the AVM below:\\[-1.5ex]

\newcommand{\compsa}{\avmlist{\avmu{VFORM & bse\\ CONT & \idx{2}}}}
\newcommand{\compsb}{\avmlist{\ }}

\noindent{}
{\avmu{PHON  & \avmd{d}{lieben, lieben, liebt, liebt} \\[-3ex]
       VFORM & \avmd{d}{bse,   bse,   fin,     fin} \\[-1ex]
       SUBJ  & \idx{1} \\
       COMPS & \avmd{d}{\compsa, \compsb, \compsb, \compsa} \\[-1ex]
       CONT  & \avmu{lieben\\ ARG1 & \idx{1}\\ ARG2 & \idx{2}} \\[-1ex]
       SLASH & \avmd{d}{\compsb, \compsa, \compsa, \compsb}
}}\\

\noindent{}%
This complex lexical entry represents the base lexical entry for the
German verb {\em lieben}, ``to love'', and the three lexical entries
that can be derived from it given the lexical rules presented in
\cite{Meurers:Minnen:96}.  The differences between these lexical
entries are encoded by the dependent disjunctions all of which are in
the same group, \(d\).  The first disjunct in each disjunction
corresponds to the base form, the second corresponds to the
application of the Complement Extraction Lexical Rule, the third
corresponds to the application of the Finitivization Lexical Rule, and
the last corresponds to the application of both rules.\footnote{These
  lexical rules are simplified versions of those presented in
  \cite{PS94}.} Modularization can be used to make this feature
structure even more efficient by splitting the group \(d\) into two
new groups
\(d_1\) and \(d_2\) as shown below.\\[-1.5ex]

\noindent{}
{\avmu{PHON  & \avmd{d_1}{lieben, liebt} \\[-3ex]
       VFORM & \avmd{d_1}{bse,    fin} \\[-1ex]
       SUBJ  & \idx{1} \\
       COMPS & \avmd{d_2}{\compsa, \compsb} \\[-1ex]
       CONT  & \avmu{lieben\\ ARG1 & \idx{1}\\ ARG2 & \idx{2}} \\[-1ex]
       SLASH & \avmd{d_2}{\compsb, \compsa}
}}\\

Another example of where modularization might prove useful is in the
treatment of typed feature structures presented in \cite{GK94}.  Their
approach produces a set of feature structures from a satisfiability
algorithm such that all of the feature structures have the same shape
but the nodes may be labeled by different types.  They then collapse
this set down to a single feature structure where nodes are labeled
with dependent disjunctions of types.  Many of the groups of
disjunctions in their feature structures can be made more efficient
via modularization.

A final example is in the {\em compaction\/} algorithm for feature
structures, presented in \cite{Gri95}.  Compaction is another
operation designed to optimize feature structures for unification.  It
takes a disjunction of feature structures, transforms them into a
single feature structure with dependent disjunctions, and then pushes
the disjunctions down in the structure as far as possible.  The result
is a large number of dependent disjunctions in the same group.  Many
of these can probably be split into new independent groups. 

\section{Contexted constraints}
\label{sec:cxcons}

\newcite{MK89} showed how a disjunction of constraints could be
replaced by an equi-satisfiable conjunction of {\em contexted
  constraints\/} as in lemma~\ref{lem:MKContextedConstraints}
below.\footnote{For a proof see \cite{MK89}.}
\begin{Lem}{Contexted Constraints}
  \label{lem:MKContextedConstraints}
  \(\phi_1\vee\phi_2\) is satisfiable
  \(\iff\) \(\bp{p\rightarrow\phi_1}\wedge\bp{\ol{p}\rightarrow\phi_2}\)
  is satisfiable, where \(p\) is a new propositional variable.
\end{Lem}
Disjunctions are replaced by conjunctions of implications from {\em
  contexts\/} (propositional formulae) to the {\em base constraints\/}
(ie. \(\phi_1\) and \(\phi_2\)).  The nature of the base constraints
is irrelevant as long as there is a satisfaction algorithm for them.
The key insight is that solving disjunctions of the base constraints
is no longer necessary since they are purely conjunctive.

Maxwell and Kaplan's goal in doing this was to have an efficient
method for {\em solving\/} disjunctive constraints.  The goal in this
paper is {\em compiling\/} disjunctive constraints into more efficient
ones for future solution.  To this end a somewhat different notion of
contexted constraint will be used as show in
lemma~\ref{lem:contextedConstraints}.
\begin{Lem}{Alternative-Case Form}
  \label{lem:contextedConstraints}
  \( \phi_1 \vee \phi_2 \) is satisfiable \(\iff\) \( \bp{a_1
    \rightarrow \phi_1} \wedge \bp{a_2 \rightarrow \phi_2} \wedge
  \bp{a_1 \vee a_2} \) is satisfiable, where \(a_1\) and \(a_2\) are
  new propositional variables.
\end{Lem}
We can see that this formulation is nearly equivalent to Maxwell and
Kaplan's by substituting \(p\) for \(a_1\) and \(\ol{p}\) for \(a_2\).
To make the formulation completely equivalent, we would need to
enforce the uniqueness of a solution by conjoining \(\ol{a}_1 \vee
\ol{a}_2\).  However, this is unnecessary since we want to permit both
solutions to be simultaneously true.  The reason for using the
modified version of contexted constraints in
lemma~\ref{lem:contextedConstraints} is that we can separate the
representation of disjunctions into a conjunction of the values that
the disjuncts can have, called the {\em alternatives}, and the way in
which the we can choose the values, called the {\em cases}.  The
alternatives are the conjunction \(\bp{a_1 \rightarrow \phi_1} \wedge
\bp{a_2 \rightarrow \phi_2}\) and the cases are the disjunction
\(\bp{a_1 \vee a_2}\).

While we could use repeated applications of
lemma~\ref{lem:contextedConstraints} to turn a disjunction of \(n\)
disjuncts into an alternative-case form, it will simplify the
exposition to have a more general way of doing this, as shown in
lemma~\ref{lem:nAryDisj}.

\newlength{\tmpa}

\newcommand{\NAryDisjLHS}{%
  \settowidth{\tmpa}{\(\:\bigvee\!\phi_i\:\)}%
  \begin{minipage}[b]{\tmpa}
    \vspace{\arrayUnStretch}
    \[\bigvee_{i\in{}N}\!\phi_i\]
  \end{minipage}
}

\newcommand{\NAryDisjRHS}{%
  \settowidth{\tmpa}{\(\:\bigwedge\!\bp{a_i\rightarrow\phi_i}\ \wedge\bigvee\!a_i\:\)}%
  \begin{minipage}[b]{\tmpa}
    \vspace{\arrayUnStretch}
    \[\bigwedge_{i\in{}N}\!\bp{a_i\rightarrow\phi_i}\ \wedge\bigvee_{i\in{}N}\!a_i\]
  \end{minipage}
}

\newlength{\arrayUnStretch}
\setlength{\arrayUnStretch}{-2em}       

\newcommand{\NAryAlts}{%
  \settowidth{\tmpa}{\(\bigwedge\!\bp{a_i\rightarrow\phi_i}\:\)}%
  \begin{minipage}[b]{\tmpa}
    \vspace{\arrayUnStretch}
    \[\bigwedge_{i\in{}N}\!\bp{a_i\rightarrow\phi_i}\]%
  \end{minipage}
}

\newcommand{\NAryCases}{%
  \settowidth{\tmpa}{\(\bigvee\!a_i\:\)}%
  \begin{minipage}[b]{\tmpa}
    \vspace{\arrayUnStretch}
    \[\bigvee_{i\in{}N}\!a_i\]
  \end{minipage}
}

\newlength{\arrayUnStretchFN}
\setlength{\arrayUnStretchFN}{-2.3ex} 

\newcommand{\NV}{%
  \settowidth{\tmpa}{\(\bigvee\!\phi_i\:\)}%
  \begin{minipage}[b]{\tmpa}
    \vspace{\arrayUnStretchFN}
    \[\bigvee_{i\in{}N}\!\phi_i\]
  \end{minipage}
}

\newcommand{\NW}{%
  \settowidth{\tmpa}{\(\bigwedge\!\phi_i\:\)}%
  \begin{minipage}[b]{\tmpa}
    \vspace{\arrayUnStretchFN}
    \[\bigwedge_{i\in{}N}\!\phi_i\]
  \end{minipage}
}

\begin{Lem}{N-ary Alternative-Case Form}
  \label{lem:nAryDisj}
  \NAryDisjLHS\ is satisfiable iff \NAryDisjRHS\ is satisfiable, where
  each \(a_i\) is a new propositional variable.\footnote{\NV\ and \NW\ 
    are disjunctions and conjunctions of formulae \(\phi_i\),
    respectively, where each \(i\) is a member of the set of
    indices, \(N\).}
\end{Lem}
Here \NAryAlts\ are the alternatives and \NAryCases\ are the cases.
So for example, \(\phi_1\vee \phi_2\vee \phi_3\vee \phi_4\) is
satisfiable just in case \(\bp{a_1\rightarrow\phi_1} \wedge
\bp{a_2\rightarrow\phi_2} \wedge \bp{a_3\rightarrow\phi_3} \wedge
\bp{a_4\rightarrow\phi_4} \wedge \bp{a_1 \vee a_2 \vee a_3 \vee a_4}\)
is satisfiable.

\section{Dependent disjunctions as contexted constraints}
\label{sec:deps}

\newcommand{\nAryByMDep}{%
  \settowidth{\tmpa}{\(\:\bigwedge\dep{d}{\phi^i_1, \phi^i_2, \cdots, \phi^i_n}\:\)}%
  \begin{minipage}[b]{\tmpa}
    \vspace{\arrayUnStretch}
    \[\bigwedge_{i\in{}M}\dep{d}{\phi^i_1, \phi^i_2, \cdots, \phi^i_n}\]
  \end{minipage}
  }

\newcommand{\AltForm}[2]{%
  \settowidth{\tmpa}{\(\:\bigwedge\bigwedge\bp{a^i_j\rightarrow\phi^i_j}\:\)}%
  \begin{minipage}[b]{\tmpa}
    \vspace{\arrayUnStretch}
    \[\bigwedge_{i\in{}#1}\bigwedge_{j\in{}#2}\bp{a^i_j\rightarrow\phi^i_j}\]
  \end{minipage}\ \ \ 
  }

\newcommand{\CaseForm}[2]{%
  \settowidth{\tmpa}{\(\:\bigvee\bigwedge{}a^i_j\:\)}%
  \begin{minipage}[b]{\tmpa}
    \vspace{\arrayUnStretch}
    \[\bigvee_{j\in{}#2}\bigwedge_{i\in{}#1}a^i_j\]
  \end{minipage}\ \ \ 
  }

The usefulness of the alternative-case form only becomes apparent when
considering dependent disjunctions.  Dependent disjunctions can be
represented by alternative-case forms as shown in
definition~\ref{def:deps} below.
\begin{Def}{Dependency Group} \label{def:deps}
  A {\em dependency group\/} is a conjunction of dependent
  disjunctions with the same name, \(d\), where each disjunction is an
  alternative-case form such that there is one alternative for every
  disjunct of every disjunction in the group, and there is one case
  for each disjunct in the group which is a conjunction of the
  alternative variables for that disjunct in every disjunction:

  \noindent{}
  \nAryByMDep \\
  \hbox{\hspace{1cm}}\(\:\equiv\:\) \AltForm{M}{N} \(\wedge\)\CaseForm{M}{N},\\
  where each \(a_j^i\) is a new propositional variable and \(N = \{1,
  2, \ldots, n\}\).
\end{Def}

So the dependent disjunction %
\dep{d}{\phi, \phi, \phi\p} \(\wedge\) \dep{d}{\psi, \psi\p, \psi\p} %
is the alternative-case form with alternatives %
\(\bp{a_1^1 \rightarrow \phi} \wedge
  \bp{a_2^1 \rightarrow \phi} \wedge
  \bp{a_3^1 \rightarrow\phi\p} \wedge
  \bp{a_1^2 \rightarrow \psi} \wedge
  \bp{a_2^2 \rightarrow \psi\p} \wedge
  \bp{a_3^2 \rightarrow \psi\p}
\) %
and cases %
\(\bp{\bp{a_1^1 \wedge a_1^2} \vee
  \bp{a_2^1 \wedge a_2^2} \vee
  \bp{a_3^1 \wedge a_3^2}}\). %
The cases enforce that the corresponding disjuncts of every disjunct
in the group must be simultaneously satisfiable.

We can now start to see where redundancy in dependent disjunctions
originates.  Because every disjunction in a group of dependent
disjunctions must have the same number of disjuncts, some of those
disjuncts may appear more than once.  In the above example for
instance, \(\phi\) occurs twice in the first disjunction and
\(\psi\p\) occurs twice in the second disjunction.  To resolve this
problem we impose the following condition, called {\em alternative
  compactness\/}: if a base constraint \(\phi_j^i\) equals another
base constraint from the same disjunction, \(\phi_k^i\), then the
alternatives variables associated with those base constraints,
\(a_j^i\) and \(a_k^i\), are also equal.\footnote{Note that this
  requires being able to determine equality of the base constraints.}
Doing this allows us to express the alternatives from the example above as %
\(\bp{a_1^1 \rightarrow \phi} \wedge
  \bp{a_3^1 \rightarrow\phi\p} \wedge
  \bp{a_1^2 \rightarrow \psi} \wedge
  \bp{a_2^2 \rightarrow \psi\p}
\), %
and the cases as %
\(\bp{\bp{a_1^1 \wedge a_1^2} \vee
      \bp{a_1^1 \wedge a_2^2} \vee
      \bp{a_2^1 \wedge a_2^2}}\).\footnote{%
      In this example, equivalent alternative variables have been
      replaced by representatives of their equivalence class.  So
      \(a_2^1\) has been replaced by \(a_1^1\) and \(a_3^2\) has been
      replaced by \(a_2^2\).}
One advantage of this is that the number of base constraints that must
be checked during satisfaction can potentially be exponentially reduced.

The next section will show how an alternative-case form for a group of
dependent disjunctions can be split into a conjunction of two (or
more) equivalent forms, thereby (potentially) exponentially reducing
the number of alternative variable interactions that must be checked
during satisfaction.

\section{Modularization}
\label{sec:modn}

Consider again the example from section~\ref{sec:depIntro}: %
\(\dep{d}{\phi, \phi, \phi\p, \phi\p} \wedge \dep{d}{\psi, \psi\p, \psi, \psi\p}\). %
Represented as a compact alternative-case form, the alternatives becomes: %
\(\bp{a_1^1 \rightarrow \phi} \wedge
  \bp{a_3^1 \rightarrow\phi\p} \wedge
  \bp{a_1^2 \rightarrow \psi} \wedge
  \bp{a_2^2 \rightarrow \psi\p}
\), %
with cases: %
\( \bp{ \bp{a_1^1 \wedge a_1^2} \vee
        \bp{a_1^1 \wedge a_2^2} \vee
        \bp{a_3^1 \wedge a_1^2} \vee
        \bp{a_3^1 \wedge a_2^2}}
\). %
The key to determining that the two disjunctions can be split into
different groups then involves determining that cases can be split
into a conjunction of two smaller cases %
\( \bp{a_1^1 \vee a_3^1} \wedge \bp{a_1^2 \vee a_2^2} \). %
If the cases can be split in this manner, we say the cases (and by
extension the group of dependent disjunctions) are {\em independent}.
\begin{Def}{Independence}
  \label{def:indep}
  A case form is {\em independent\/} \(\iff\) it is equivalent to the
  conjunction of 2 (or more) other cases forms:\\[-2ex]
  
  \noindent{}
  \CaseForm{M}{N} \(\equiv\) \CaseForm{M\p}{N\p}\ \ \  \(\wedge\) \CaseForm{M\pp}{N\pp}\\
where \(M\p\) and \(M\pp\) partition \(M\).
\end{Def}
So in the above example, \(M = \{1,2\}\) where 1 represents the first
disjunction and 2 represents the second.  That makes \(M\p = \{1\}\)
and \(M\pp = \{2\}\).  While \(M\p\) and \(M\pp\) are derived from
\(M\), the elements of the \(N\)s are arbitrary.  But a consequence of
definition~\ref{def:indep} is that \(\card{N} = \card{N\p} \times
\card{N\pp}\).  This will be proved in section~\ref{sec:freeCombElim}.
The size of the \(N\)s, however, represent the number of cases.  So
for instance in the above example, \(N\) might equal \(\{1,2,3,4\}\)
since there are four disjuncts in the original case form, while
\(N\p\) might equal \(\{1,2\}\) and \(N\pp\), \(\{1,2\}\), since the
smaller case forms each contain two disjuncts.

The process of splitting a group of dependent disjunctions into
smaller groups is called {\em modularization}.  Modularizing a group
of dependent disjunctions amounts to finding a conjunction of case
forms that is equivalent to the original case form.  The
modularization algorithm consists of two main steps.  The first is to
take the original case form and to construct a pair of possibly
independent case forms from it.  The second step is to check if these
case forms are actually independent from each other with respect to
the original one.  The modularization algorithm performs both of these
steps repeatedly until either a pair of independent case forms is
found or until all possible pairs have been checked.  If the later,
then we know that the original dependent disjunction is already
modular.  If on the other hand we can split the case forms into a pair
of smaller, independent case forms, then we can again try to
modularize each of those, until all groups are modular.

To construct a pair of potentially independent case forms, we first
need to partition the set of alternative variables from the original
case form into two sets.  The first subset contains all of and only
the variables corresponding to some subset of the original disjunctions
and the second subset of variables is the complement of the first,
corresponding to all of and only the other disjunctions.  From these
subsets of variables, we construct two new case forms from the original
using the operation of {\em confinement}, defined below.
\begin{Def}{Confinement}
  \(\conf\bp{\CaseForm{M}{N}, M\p}\)\\ is the {\em confinement\/} of
  \CaseForm{M}{N} with respect to a set of indices, \(M\p\),\\[.5ex] \(\iff\)
  \(\conf\bp{\CaseForm{M}{N}, M\p} \equiv \dnf\bp{\CaseForm{M\p}{N}\:\:}\),\\
  where \(M\p \subseteq M\).
\end{Def}
Constructing the confinement of a case form is essentially just
throwing out all of the alternative variables that are not in \(M\p\).
However, only doing this might leave us with duplicate disjuncts, so
converting the result to DNF removes any such duplicates.

To make the definition of confinement clearer, consider the following
conjunction of dependent disjunctions: %

\( \dep{d}{\phi, \phi,   \phi,   \phi,   \phi\p, \phi\p} \wedge
   \dep{d}{\psi, \psi\p, \psi,   \psi\p, \psi,   \psi\p} \wedge \)

\(\dep{d}{\chi, \chi,   \chi\p, \chi\p, \chi\p, \chi\p}
\). %

\noindent
This is equivalent to the compact alternative form:\footnote{In this
  example, equivalent alternative variables have again been replaced
  by representatives of their equivalence class.  So for instance,
  \(a_2^1\), \(a_3^1\) and \(a_4^1\) are all represented by
  \(a_1^1\).}

\(\bp{a_1^1 \rightarrow\phi} \wedge
  \bp{a_5^1 \rightarrow\phi\p} \wedge
  \bp{a_1^2 \rightarrow\psi} \wedge \)

\(\bp{a_2^2 \rightarrow\psi\p} \wedge
  \bp{a_1^3 \rightarrow\chi} \wedge
  \bp{a_3^3 \rightarrow\chi\p}
\), %

\noindent
and the following case form: \(case =\)%

\( \bp{\bp{a_1^1 \wedge a_1^2 \wedge a_1^3} \vee
       \bp{a_1^1 \wedge a_2^2 \wedge a_1^3} \vee
       \bp{a_1^1 \wedge a_1^2 \wedge a_3^3} \vee \)

\(     \bp{a_1^1 \wedge a_2^2 \wedge a_3^3} \vee
       \bp{a_5^1 \wedge a_1^2 \wedge a_3^3} \vee
       \bp{a_5^1 \wedge a_2^2 \wedge a_3^3}}
\). %

\noindent
Now we can compute the confinements.  For instance, %

\( \conf\bp{case,\{1,2\}} = \dnf
   \bp{\bp{a_1^1 \wedge a_1^2} \vee
       \bp{a_1^1 \wedge a_2^2} \vee \)

\(     \bp{a_1^1 \wedge a_1^2} \vee
       \bp{a_1^1 \wedge a_2^2} \vee
       \bp{a_5^1 \wedge a_1^2} \vee
       \bp{a_5^1 \wedge a_2^2}}
\). %

\noindent
After removing duplicates we get: %

\( \conf\bp{case,\{1,2\}} =\)

\(   \bp{\bp{a_1^1 \wedge a_1^2} \vee
       \bp{a_1^1 \wedge a_2^2} \vee
       \bp{a_5^1 \wedge a_1^2} \vee
       \bp{a_5^1 \wedge a_2^2}}
\). %

\noindent
Likewise, for the complement of \(M\p\) with respect to \(M\), we get: %

\( \conf\bp{case,\{3\}} =
   \bp{\bp{a_1^3} \vee
       \bp{a_3^3}}
\). %

Now we just need to test whether two confined case forms are
independent with respect to the original.  This is done with the {\em
  free combination\/} operation, shown in
definition~\ref{def:freeComb}.

\begin{Def}{Free Combination \(\otimes\)}
  \label{def:freeComb}
  The {\em free combination\/} of two case forms is the disjunctive
  normal form of their conjunction:

  \(\Case\p \otimes \Case\pp \equiv \dnf\bp{\Case\p \wedge \Case\pp} \)
\end{Def}
The two case forms, \(\Case\p\) and \(\Case\pp\), are DNF formulae.
To compute the free combination, we conjoin them and convert the
result back into DNF\@.  They are independence if their free
combination is equal to the original case form, \(\Case\).

For example, the free combination of the two confinements from above, %

\( \bp{\bp{a_1^1 \wedge a_1^2} \vee
       \bp{a_1^1 \wedge a_2^2} \vee
       \bp{a_5^1 \wedge a_1^2} \vee
       \bp{a_5^1 \wedge a_2^2}}
\) %
and

\( \bp{\bp{a_1^3} \vee
       \bp{a_3^3}}
\) %

\noindent
is %

\( (  \bp{a_1^1 \wedge a_1^2 \wedge a_1^3} \vee
      \bp{a_1^1 \wedge a_2^2 \wedge a_1^3} \vee
      \bp{a_5^1 \wedge a_1^2 \wedge a_1^3} \vee \)

\(    \bp{a_5^1 \wedge a_2^2 \wedge a_1^3} \vee
      \bp{a_1^1 \wedge a_1^2 \wedge a_3^3} \vee
      \bp{a_1^1 \wedge a_2^2 \wedge a_3^3} \vee \)

\(    \bp{a_5^1 \wedge a_1^2 \wedge a_3^3} \vee
      \bp{a_5^1 \wedge a_2^2 \wedge a_3^3}) \) %

\noindent
which is not equal to the original case form: %

\( \bp{\bp{a_1^1 \wedge a_1^2 \wedge a_1^3} \vee
       \bp{a_1^1 \wedge a_2^2 \wedge a_1^3} \vee
       \bp{a_1^1 \wedge a_1^2 \wedge a_3^3} \vee \)

\(     \bp{a_1^1 \wedge a_2^2 \wedge a_3^3} \vee
       \bp{a_5^1 \wedge a_1^2 \wedge a_3^3} \vee
       \bp{a_5^1 \wedge a_2^2 \wedge a_3^3}}
\), %

\noindent%
so the first two disjunctions are not independent from the third.
However, the second disjunction is independent from the first and the
third since \(\conf\bp{case,\{2\}} =  \bp{\bp{a_1^2} \vee \bp{a_2^2}}\),
and \(\conf\bp{case,\{1,3\}} = \)
\( \bp{\bp{a_1^1 \wedge a_1^3} \vee
       \bp{a_1^1 \wedge a_3^3} \vee
       \bp{a_5^1 \wedge a_3^3}}
\), %
and their free combination is equal to the original case form.  Therefore,
the original formula is equivalent to
\( \dep{d\p}{\psi, \psi\p} \wedge
   \dep{d\pp}{\phi, \phi, \phi\p} \wedge
   \dep{d\pp}{\chi, \chi\p, \chi\p}
\). %

\section{Free combination elimination}
\label{sec:freeCombElim}

The last section showed an effective algorithm for modularizing groups
of dependent disjunctions.  However, even though this is a compile
time algorithm we should be concerned about its efficiency since it
has exponential complexity.  The main source of complexity is that we
might have to check every pair of subsets of disjunctions from the
group.  In the worst case this is unavoidable (although we do not
expect natural language grammars to exhibit such behavior).  Other
sources of complexity are computing the free combination and testing
the result against the original case form.  Luckily it is possible to
avoid both of these operations.  This can be done by noting that both
the original case form and each of the confined case forms are in DNF\@.
Therefore it is a necessary condition that if the free combination of
the confinements is the same as the original case form then the
product of the number of disjuncts in each confinement,
\(\card{\Case\p} \times \card{\Case\pp}\), must equal the number of
disjuncts in the original case form, \(\card{\Case}\).  Moreover,
since both confinements are derived from the original case form, it is
also a sufficient condition.  This is shown more formally in
theorem~\ref{thm:freeCombElim}.

\begin{Thm}{Free combination elimination}
  \label{thm:freeCombElim}
  \( \Case = \Case\p \otimes \Case\pp \Longleftrightarrow \card{\Case}
  = \card{\Case\p} \times \card{\Case\pp} \)
\end{Thm}

\paragraph{Proof \(\Longrightarrow\)}

We assume that \(\Case\p \otimes \Case\pp = \Case\).  Since both
\(\Case\p \otimes \Case\pp\) and \(\Case\) are in \(\DNF\) and
\(\DNF\) is unique, we know that \(\card{\Case\p \otimes \Case\pp} =
\card{\Case}\).  We also know that \(\Case\p\) and \(\Case\pp\) have
no disjuncts in common because they have no alternative variables in
common, so \(\card{\Case\p \otimes \Case\pp} = \card{\Case\p} \times
\card{\Case\pp}\).  Therefore, \(\card{\Case} = \card{\Case\p} \times
\card{\Case\pp}\). \qed

\paragraph{Proof \(\Longleftarrow\)}
Again since \(\Case\p\) and \(\Case\pp\) have no disjuncts in common,
we know that \(\card{\Case\p \otimes \Case\pp} = \card{\Case\p} \times
\card{\Case\pp}\) and therefore, that \(\card{\Case} = \card{\Case\p
  \otimes \Case\pp}\).  Every disjunct in \(\Case\) can be represented
as \(A\p\wedge A\pp\) where \(A\p\) is a disjunct in \(\Case\p\) and
\(A\pp\) is a disjunct in \(\Case\pp\).  So the disjuncts in \(\Case\p
\otimes \Case\pp\) must be every conjunction of possible \(A\p\)s and
\(A\pp\)s.  So \(\Case\p \otimes \Case\pp\) must contain all of the
disjuncts in \(\Case\) and it could contain even more, but then
\(\card{\Case\p \otimes \Case\pp} > \card{\Case}\).  However, since
\(\card{\Case} = \card{\Case\p \otimes \Case\pp}\), \(\Case\p \otimes
\Case\pp\) must contain exactly the disjuncts in \(\Case\) and
therefore \(\Case = \Case\p \otimes \Case\pp\). \qed

We can see that this would have helped us in the previous example to
know that \(\conf\bp{case,\{1,2\}}\) could not be independent from
\(\conf\bp{case,\{3\}}\) with respect to \(\Case\) because
\(\card{\conf\bp{case,\{1,2\}}} = 4\) and
\(\card{\conf\bp{case,\{3\}}} = 2\) but \(\card{\Case} = 6\), not
\(8\).  Conversely, since \(\card{\conf\bp{case,\{1,3\}}} = 3\) and
\(\card{\conf\bp{case,\{2\}}} = 2\), we know immediately that these
case forms are independent.

This theorem also allows us to perform other combinatorial short cuts,
such as noting that if the number of disjuncts in the original case
form is prime then it is already modular.

\section{Conclusion}
\label{sec:concl}

This paper has presented an efficient form for representing dependent
disjunctions and an algorithm for detecting and eliminating redundant
interactions within a group of dependent disjunctions.  This method
should be useful for any system which employs dependent disjunctions
since it can eliminate exponential amounts of processing during
constraint satisfaction.


\end{document}